\documentclass[runningheads]{llncs}

\usepackage{eccv}

\usepackage{diagbox}
\usepackage{makecell}
\usepackage{xspace}
\usepackage{subcaption}
\usepackage{multirow}
\usepackage{wrapfig}
\usepackage{pifont}
\usepackage{xcolor,colortbl,pgf}
\usepackage{eccvabbrv}

\usepackage{graphicx}
\usepackage{booktabs}

\usepackage[accsupp]{axessibility}  % Improves PDF readability for those with disabilities.

\usepackage[pagebackref,breaklinks,colorlinks,citecolor=eccvblue]{hyperref}
\usepackage{orcidlink}

\begin{document}

% \title{Generating Weights from Diffusion Model For Fast Model Training} 
\title{Efficient Training with Denoised Neural Weights
% Generating Weights from Diffusion Model For 
} 
\titlerunning{Efficient Training with Denoised Neural Weights}

\author{Yifan Gong\inst{1,2}\thanks{Work done during internship at Snap Inc.}\orcidlink{0000-0002-3912-097X}  \and
Zheng Zhan\inst{2}\orcidlink{0000-0002-3882-5484} \and
Yanyu Li\inst{1,2}\orcidlink{0000-0003-1240-4785} \and
Yerlan Idelbayev\inst{1}\orcidlink{0000-0002-0179-467X} \and
Andrey Zharkov\inst{1} \orcidlink{0000-0001-9662-4552} \and
Kfir Aberman\inst{1} \orcidlink{0000-0002-4958-601X
} \and Sergey Tulyakov\inst{1} \orcidlink{0000-0003-3465-1592} \and
Yanzhi Wang\inst{2} \orcidlink{0000-0002-4325-521X} \and
Jian Ren\inst{1} \orcidlink{0000-0002-0511-7473}
}

\authorrunning{Y. Gong, Z. Zhan, et al.}

\institute{$^1$ Snap Inc. $^2$ Northeastern University\\
Project Page: {\url{https://yifanfanfanfan.github.io/denoised-weights/}}}

\maketitle

\begin{abstract}
Good weight initialization serves as an effective measure to reduce the training cost of a deep neural network (DNN) model. The choice of how to initialize parameters is challenging and may require manual tuning, which can be time-consuming and prone to human error. To overcome such limitations, this work takes a novel step towards building a \emph{weight generator} to synthesize the neural weights for initialization. We use the image-to-image translation task with generative adversarial networks (GANs) as an example due to the ease of collecting model weights spanning a wide range. Specifically, we first collect a dataset with various image editing concepts and their corresponding trained weights, which are later used for the training of the weight generator. To address the different characteristics among layers and the substantial number of weights to be predicted, we divide the weights into equal-sized blocks and assign each block an index. Subsequently, a diffusion model is trained with such a dataset using both text conditions of the concept and the block indexes. By initializing the image translation model with the denoised weights predicted by our diffusion model, the training requires only $43.3$ seconds. Compared to training from scratch (\ie, Pix2pix), we achieve a $15\times$ training time acceleration for a new concept while obtaining even better image generation quality. 
% We will release our dataset, code, and the pre-trained weight generator.

% Inspired by the recent advances in diffusion models with promising image generation ability, we study the possibilty of build a weight generator to help generate weight initialization for different tasks. 
\end{abstract}

\section{Introduction}
Efficient training for deep neural networks (DNN) not only accelerates the model development process but also reduces the requirements for computational resources and costs. Many prior works have investigated efficient training strategies, such as sparse training~\cite{lee2018snip,tanaka2020pruning,wang2020picking,bellec2017deep,dettmers2019sparse,evci2020rigging,yuan2021mest,kong2023peeling,wang2022sparcl} and low-bit training~\cite{sun2020ultra,venkataramani2021rapid,wortsman2024stable}. However, achieving efficient training is often hindered by challenges in initializing model weights effectively. While some efforts have been conducted in the domain of weight initialization~\cite{zhang2019fixup,bachlechner2021rezero,de2020batch,huang2020improving,cordonnier2019relationship,d2021convit}, determining the appropriate schemes to use across different tasks remains challenging. Tuning parameters for weight initialization can be time-consuming and prone to human error, leading to sub-optimal performance and increased training time.

To tackle such challenges, inspired by recent advances in designing HyperNetworks~\cite{ruiz2023hyperdreambooth,erkocc2023hyperdiffusion,wang2024neural}, for the first time, we investigate the feasibility of building a \emph{weight generator} to provide better weight initialization across different tasks, thus reducing the training time and resource consumption for obtaining a well-trained DNN model.
We use image-to-image translation tasks trained with GAN models~\cite{isola2017image,zhu2017unpaired,park2020contrastive} as an example to unfold our design for predicting neural weights, though our framework is a general design that is not restricted to generating GAN weights. The reason for the choice is the easy acquisition of a vast volume of different weights trained on different datasets.

More specifically, the weight generator can predict the initialized weight for unseen new concepts and styles. To reduce the number of weights to be predicted, we apply Low-Rank Adaptation (LoRA)~\cite{hu2021lora} to the image generation model, resulting in many fewer model parameters while still maintaining high-quality image generation.
% sufficient to transfer the generative domain of the GAN models while containing many fewer parameter numbers. 
As the GAN model is composed of different types of layers with different sizes and number of weights, we group the weights and divide them into equal-sized blocks. A diffusion process~\cite{sohl2015deep,song2019generative,song2020score,ldm} is leveraged to model the space of well-trained weights from GAN models. Hence, the weights estimation by training diffusion model, namely, the weight generator, becomes possible.
% weight initialization space of GAN models. 
To improve the performance of the weight generator, we further incorporate the block index as a further conditioning mechanism in the weight generator, by adopting a sinusoidal positional encoding scheme and  computing the embeddings for block indexes. The embedding provides the weight generator with information about the position of each weight block within all the model weights. After obtaining the weight generator, to train a GAN-based image translation model, we conduct a fast inference of the weight generator through a single-step denoising process and use the predicted weights to initialize the GAN model. The GAN model only requires a subsequent efficient fine-tuning process to obtain high-quality image generation results, significantly reducing the time consumption for obtaining the model for the newly unseen concept. We summarize our contributions as follows:

\begin{itemize}
    \item We propose a framework to generate weight initializations across different concepts/styles to efficiently train the GAN models for the image translation.
    \item We collect a vast ground-truth dataset of LoRA weights for different concepts/styles with the help of diffusion models (\ie, preparing paired image datasets), which serves as the foundation for weight generator training.
    \item We introduce an efficient design for a weight generator by utilizing a diffusion process, which incorporates both textual concept information and block indexes as inputs. To deal with different layer types and weight shapes, we organize weights into equal-sized one-dimensional blocks, significantly reducing computational overhead. These block indexes are seamlessly integrated into the weight generator design by combining them with time step embeddings. Therefore, the weight generator has the information about the position of each weight block within all the model weights.
    \item Our proposed framework can predict the initialized neural weights of a GAN model with a single denoising step, taking only $1.19$ seconds. By initializing with the predicted weights, a fast fine-tuning process can convey the target style in $42.1$ seconds. Compared to training from scratch (\ie, Pix2pix~\cite{isola2017image}), we reduce the total training time by $15\times$, while maintaining even \emph{better} image generation quality. Compared to other efficient training method~\cite{gong20242}, we can save the training time by $4.6 \times$.
    % reducing the total time consumption for a new concept by at least 4.6$\times$ than baseline methods.
\end{itemize}

\section{Related Work}

\subsection{Efficient Training}
Efficient training of DNNs has been a central point in machine learning research, aiming to reduce computational costs and memory requirements during model training while maintaining model performance. Sparse training methods \cite{lee2018snip,tanaka2020pruning,wang2020picking,bellec2017deep,dettmers2019sparse,evci2020rigging,yuan2021mest,kong2023peeling,wang2022sparcl} explore faster DNN training by applying sparse masks to the model. Static sparse training \cite{lee2018snip,tanaka2020pruning,wang2020picking} executes
traditional training after first pruning the model with a fixed sparse mask, which typically results in lower accuracy and higher computation and memory consumption for the pruning stage. On the contrary, dynamic mask methods \cite{bellec2017deep,dettmers2019sparse,evci2020rigging} start with a sparse model structure from an untrained dense model and then combine sparse topology exploration with the sparse model training, which adjusts the sparsity topology during training while maintaining a low memory footprint. Besides applying sparse masks on weights and gradients, some recent works \cite{yuan2021mest,kong2023peeling,wang2022sparcl} also investigate incorporating data efficiency with different data selection approaches for better training accelerations. Meanwhile, another direction of the research explores low-bit training of DNNs to pursue model training efficiency \cite{sun2020ultra,venkataramani2021rapid,wortsman2024stable}. However, using lower precision typically leads to an accuracy drop. A good weight initialization is essential to stabilize training, enable a higher learning rate,
accelerate convergence, and improve generalization. Existing works explore rescaling paradigms \cite{zhang2019fixup,bachlechner2021rezero,de2020batch,huang2020improving} or leverage the relationship between layers \cite{cordonnier2019relationship,d2021convit}. However, determining schemes to use is still a challenging task and prone to human errors for different tasks. Meanwhile, LoRA methods \cite{hu2021lora,dettmers2024qlora} aim to exploit the inherent low-rank structure present in DNN weight matrices to reduce computational complexity and memory requirements for fine-tuning from a pre-trained model on a specific, smaller dataset to specialize its performance on a particular task or domain. By keeping the original model unchanged and adds small, changeable parts to each layer of the model, LoRA methods can significantly reduce the number of parameters and operations required for forward and backward passes, which serves as an effective complement in the efficient training direction.
\subsection{HyperNetwork}
HyperNetworks have emerged as a promising approach in the field of generative AI by generating model parameters. HyperDreamBooth \cite{ruiz2023hyperdreambooth} introduces a HyperNetwork capable of generating personalized weights from a single image. By leveraging these weights within the diffusion model, HyperDreamBooth enables the generation of personalized faces with high subject details in diverse styles and contexts followed by a fast fine-tuning process. HyperDiffusion \cite{erkocc2023hyperdiffusion} operates directly on MLP weight instead of directly applying generative modeling on implicit neural fields. It first collects a dataset of neural field multilayer perceptrons (MLPs) overfitting on 3D or 4D shapes. The dataset is then used to train the HyperNetwork, which is an unconditional generative model, to predict the MLP weights for 3D or 4D shape generation. Neural Network Diffusion \cite{wang2024neural} works on generating model weights for image classification tasks. However, the weight generation is based on first collecting multiple trained model weights for specific model architecture on the target dataset. Furthermore, it only works on generating the weights for two normalization layers.   
% As the domain changes for the personalization task is not huge, 
% Hyperdreambooth, 3D

% Can we train more effectively?

% Initialization generating + fast finetune

% Possible apply to image classification 

% We use GAN as it's easier to collect data

\section{Motivations and Challenges}

Effective weight initialization is crucial for stabilizing training, facilitating a faster learning rate, expediting convergence, and enhancing generalization ability. However, identifying good weight initializations across different tasks remains challenging. Inspired by recent advances in HyperNetwork, we hope to investigate whether we can build a \emph{weight generator} to obtain good weight initialization, thus reducing training time and resource consumption. Unlike the popular image/video generation, little research effort has been paid to explore weight generation. Building such a weight generator is promising yet challenging. The first significant challenge comes from the different layer types within DNN architectures. The weights in each layer exhibit diverse sizes and shapes, necessitating a weight generation approach capable of accommodating this heterogeneity. Second, the weight generator must possess the capacity to generate a substantial number of parameters efficiently, ensuring comprehensive coverage across the network. Third, the inference of the weight generator should be fast and efficient to save time in obtaining the weights for a new task. Addressing these challenges holds promise for building better DNN training paradigms with higher efficiency and effectiveness of deep learning systems. Thus, in this work, we study the construction of the weight generator for better weight initializations. We aim to show the generation ability not only restricted to the weight initialization for a single model architecture on a certain dataset, such as ResNet-18 on CIFAR-10 as in \cite{wang2024neural}, but across the models for different tasks. To achieve this, we take the generation of initialization weights for GANs for image-to-image translation tasks as an example to show our methods due to the ease of collecting diverse datasets for the GAN models. Our method is not restricted to the GAN architecture or the image-to-image translation task. 
\section{Method}

% \textcolor{red}{Framework Overview, a network to predict weights, fine tune to obtain on-device model}
% \textcolor{red}{Each Component figure}
% \subsection{Framework Overview}
\begin{figure}[t]
     \centering
     \includegraphics[width=0.8\textwidth]{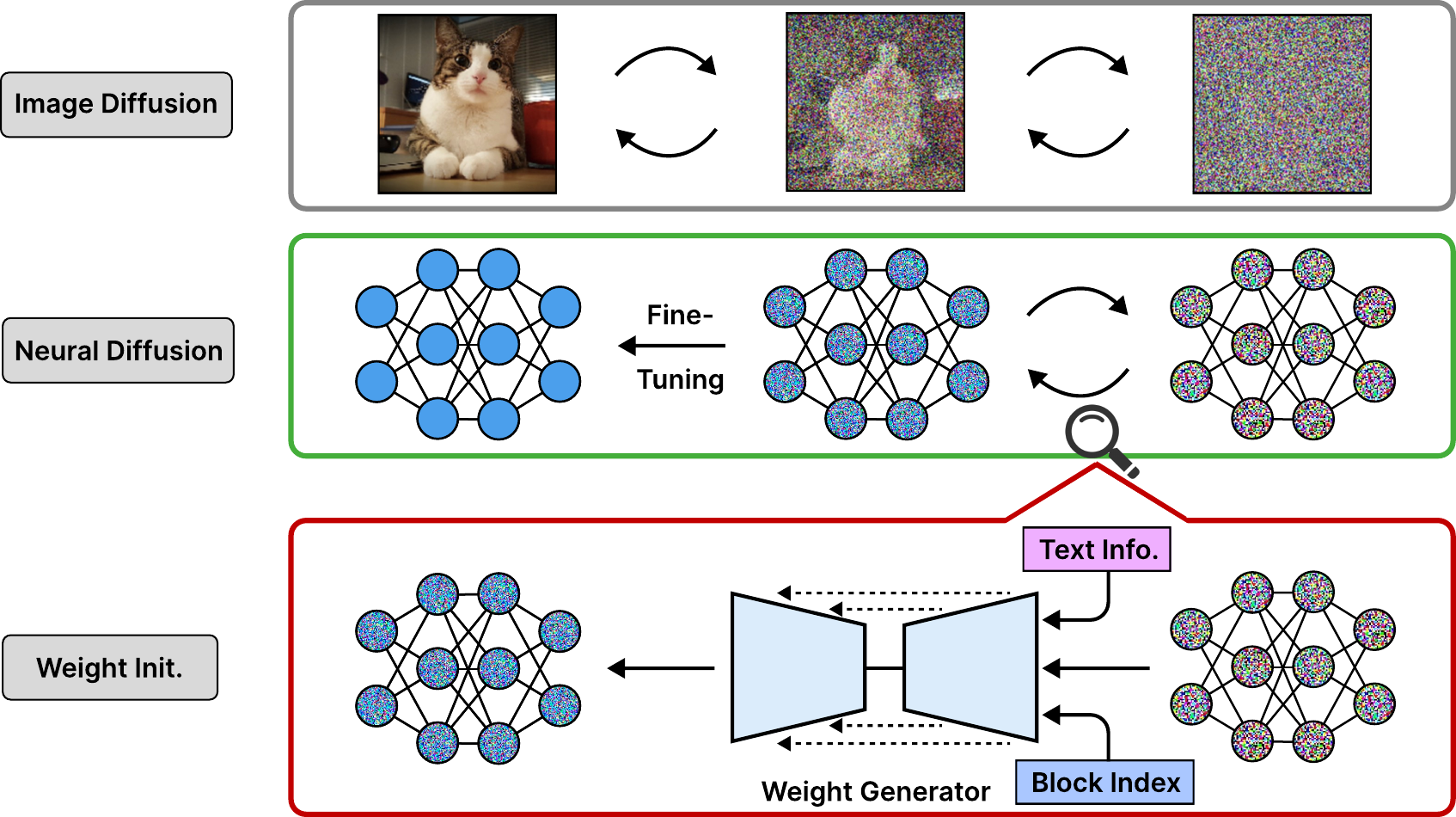}  
     \caption{The \textbf{framework overview} of our weight generator design. The standard diffusion process turns an image into noise in the forward pass and reverses a clean image from pure noise in the reverse process. Our weight generator is designed to turn a noise to weight initializations for efficient training purposes. Given the text information and block index, the weight generator provides the corresponding weight values.  }
    \label{fig:generator_framework}  
       \vspace{-1.2em}
\end{figure}

Our objective is to train a weight generator to predict the weight initializations for different tasks. We take GANs for image-to-image translation tasks as an example to demonstrate the effectiveness of our method. When there is a new concept/style, we can query the weight generator to provide the weight values for the initialization. The weight generator is modeled with a diffusion process, as illustrated in Fig. \ref{fig:generator_framework}. Different from image diffusion models that reverse a clean image from a purse noise, our framework targets turning the noise into weight values used for the initialization. By plugging in the predicted weight values, a fast fine-tuning process is conducted to achieve the efficient training of the GAN models for the target style. The core of our framework is the design of the weight generator. To build this weight generator, we elaborate on how to create the training dataset for the weight generator in Sec. \ref{sec:data_collection}, the data format for the training and inference of the weight generator in Sec. \ref{sec:data_format}, the architecture and training objective of the weight generator in Sec. \ref{sec:training_obj}, and the fine-tuning process after the weight prediction is Sec. \ref{sec:finetune}.

\subsection{Dataset Collection} \label{sec:data_collection}
In order to effectively train a weight generator for generating weight initializations of GAN models across various concepts, we need to collect a large-scale ground-truth weight value dataset for different concepts. To obtain the ground-truth weight value dataset, a large-scale prompt dataset becomes crucial. By using the concepts/styles in the prompt dataset, we can achieve image collection with diffusion models to obtain a substantial collection of images representative of each target concept. The images for each concept/style are further leveraged to train the GANs for the obtaining of the ground-truth GAN weights. 

As the foundation of data preparation for weight generator training, the prompt dataset should include diverse visual concepts/styles to enable the weight generator to learn comprehensive representations for initializing GANs tailored to specific tasks. However, the process of collecting such a dataset poses great challenges. Ensuring diversity and representativeness across different concepts/styles demands considerable data. Moreover, the collected prompts are further used to generate images in the target concept/style with diffusion models. 

To construct our prompt dataset for training a reliable weight generator for GAN weight initialization, we adopt a systematic approach that integrates both large language models (LLMs) for style \textbf{generation} and \textbf{augmentation} to ensure richness and diversity in conceptual representation. We begin by sketching out three broad categories: 1) art concepts, 2) characteristic concepts, and 3) facial modification concepts. Within each category, we leverage a large language model (ChatGPT-3.5 \cite{brown2020language} ) to ask it to generate a spectrum of textual descriptions encompassing various concepts. By filtering out redundant concepts/styles, we further conduct an augmentation method by querying a large language model (Vicuna \cite{chiang2023vicuna}) to provide concepts/styles with similar meanings but different representations. To further enrich the prompt dataset, we permute and combine concepts/styles across different categories. Through the process, we are able to curate a large-scale prompt dataset that not only spans diverse conceptual domains but also captures intricate stylistic differences, providing the foundation for the training of the weight generator for better weight initialization. 

After the prompt dataset collection,  we use the diffusion models to edit real images to obtain the edited images for each concept/style in the prompt dataset, forming pairs of data for GAN training. Here, we adopt a generator with a hybrid of ResNet blocks and transformer blocks as in paper \cite{gong20242} due to the effectiveness of the model and the hybrid architecture design to show the generation ability of our method on different types of layers. Following the GAN training process, we build a dataset of weights from GAN checkpoints for different concepts/styles. To further augment the weight value dataset, we save $K$ checkpoints through the training process for each concept/style after the FID performance converges.
% employ style augmentation methodologies to expand upon these initial textual descriptions, thereby encapsulating a broader range of stylistic variations within each concept category. Furthermore, to promote cross-pollination of ideas and styles, we systematically permute and combine concepts and styles across different categories, fostering a rich and interconnected dataset landscape. This meticulous process enables us to curate a large-scale prompt dataset that not only spans diverse conceptual domains but also captures intricate stylistic nuances, laying a solid foundation for training our weight generator for GAN initialization.

\subsection{Data Format Design for Weight Generator} \label{sec:data_format}
To train a weight generator capable of efficiently producing weight initializations for GAN models across diverse concepts, it is important to design the weight format for both training and inference. The objective is whenever a new concept is provided as the input to the weight generator, it can generate the weight initialization of all layers for the concept. Given there exist multiple different types of layers within the model such as fully connected (FC), convolutional (CONV), and batch normalization (BN) layers, and the varying sizes and dimensions across the layers, designing the appropriate data format becomes crucial and challenging. Furthermore, the scale of weights in a GAN model is typically on the scale of millions, posing more challenges to the data format design. 

A larger amount of weights to be predicted leads to more difficulties for the weight generator. To alleviate this, we apply Low-Rank Adaptation (LoRA) \cite{hu2021lora} to different layers to greatly reduce the number of weights to be predicted. For instance, for a CONV layer $i$ with weights $\mathbf{w}_i \in \mathbb{R}^{c\times f \times
k_h\times k_w}$,  we apply two low-rank matrices with rank $r_i$, \ie $ \mathbf{w}_{i}^A \in \mathbb{R}^{c\times r_i \times k_h \times k_w}$ as LoRA down layer, and  $ \mathbf{w}_{i}^B \in \mathbb{R}^{r_i \times f \times 1\times1}$ as LoRA up layer, to approximate the weight change. By doing so, the total amount of weights to be predicted is reduced from 7.06M to 0.22M. 
{We show that finetuning LoRA weights are sufficient to transfer the generative domain of the GAN model.}
Though greatly reducing the weight number, directly predicting all 0.22M weights simultaneously through inference of the weight generator once is still challenging. It requires a large weight generator with huge computation and memory burdens. 

To tackle this, we partition the weights into groups to mitigate the computational complexity and enhance the feasibility of fitting the weight generator into memory during both training and inference. As different layers have different statistical characteristics, we group the LoRA down and up layers for each layer $i$, with the associated BN layers if applicable, into one group. Still, each group has a different number of weights and shapes. Thus, we further flatten the weights into  1-dimensional vectors and divide the weights into $N$ equal-sized blocks, each with $b$ weights. Thus, the data format is denoted as $<n, \mathbf{w}_n, T>$, where $n$ is the block index, $\mathbf{w}_n \in \mathbb{R}^b$ is the flattened 1-dimensional weight vector for the $n$-th weight block, and $T$ denotes the text prompt of current concept/style. The advantages of using such a data format include: 1) works for different types and shapes of layers; 2) reduces the computation complexity and difficulty for prediction; and 3) makes the weight generator easier to fit into memory.

\subsection{Weight Generator Training} \label{sec:training_obj}
\begin{wrapfigure}{r}{0.6\textwidth}
\vspace{-8mm}
\centering{
\begin{tabular}{c}
\includegraphics[width=0.6\textwidth]{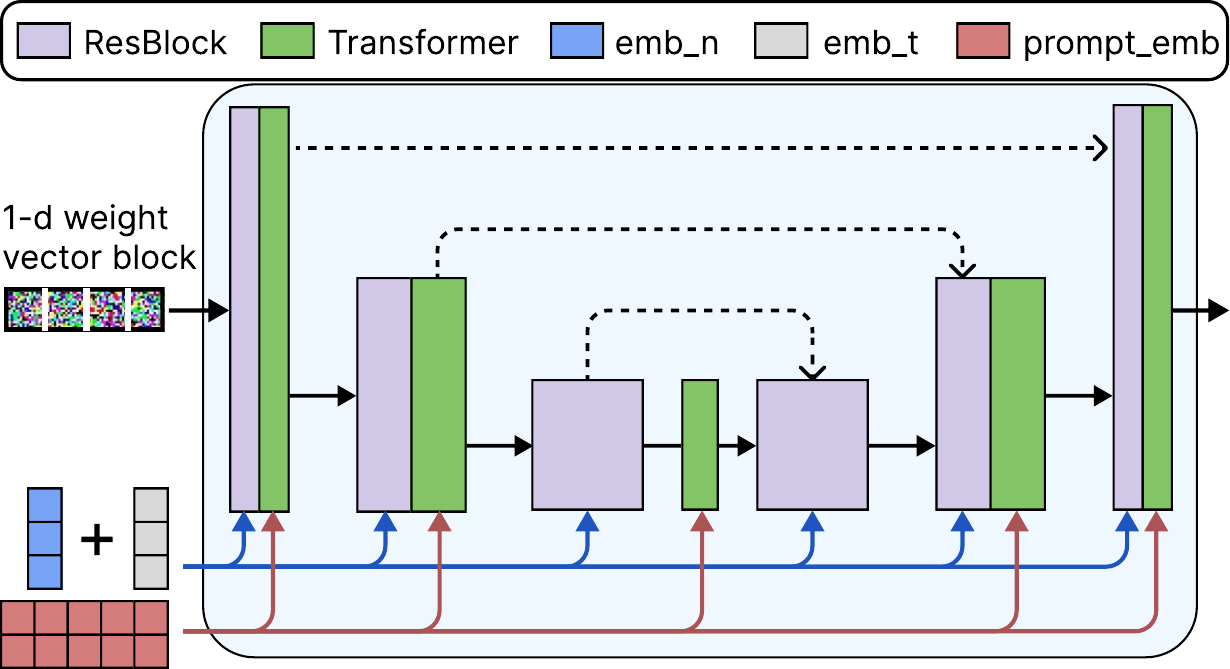}
\end{tabular}
}

\caption{\textbf{The UNet Weight Generator}. The weight generator is composed of 1-d ResBlocks and 1-d Transformer blocks. The block embedding $emb_n$ is combined with the time step embedding $emd_t$ to be leveraged in each ResBlock. 
}
\vspace{-8mm}
\label{fig: unet_arch}
\end{wrapfigure}
Using our dataset of weight values, we train a generative model that learns to provide the weight initializations for other concepts/styles.
% Specifically, $\mathbf{\epsilon}$ predicts the distribution of weight initializations $p_\mathbf{\epsilon}(\mathbf{w}_n^*|\mathbf{w}_n, n, T)$, where $\mathbf{w}_n$ is the random initialization for the weights in the $n$-th block.
We model the weight initialization space of GANs through a diffusion process. The generator is a UNet weight information creator $\hat{\mathbf{\epsilon}}_\theta$ parameterized by $\theta$ for 1-dimensional vectors, which is demonstrated in Fig. \ref{fig: unet_arch}. We diffuse the weight block $\mathbf{w}_n$ from a real weight distribution $p(\mathbf{w}_n)$ into a noisy version and train the denoising UNet to gradually reverse this process, generating weights from Gaussian noise.  The training can be formulated as the following noise prediction problem:
\begin{equation} \label{eq:diff}
    \min_\theta \mathbb{E}[\| \hat{\epsilon}_\theta(\mathbf{w}_n^t,t,n,\tau(T)) - \mathbf{\epsilon} \|_2^2],
\end{equation}
where $t$ refers to the time step; $\epsilon$ is the ground-truth noise; $\mathbf{w}_n^t = \alpha_t \mathbf{w}_n + \sigma_t \epsilon$ is the noisy weight for block $n$;  $\alpha_t$ and $\sigma_t$ are the strengths of signal and noise, respectively, decided by a noise scheduler; $\tau$ is a frozen text encoder such as CLIP \cite{radford2021learning}. To incorporate the block index as a further conditioning mechanism in our weight generator, we adopt a strategy inspired by the sinusoidal positional encoding commonly used in sequence-to-sequence models \cite{vaswani2017attention}. We compute a sinusoidal block index encoding, which serves to provide the weight generator with information about the position of each weight block within all the model weights.  Specifically, let $N$ denote the total number of weight blocks and $d$ denote the dimensionality of the encoding. The sinusoidal block index encoding $\text{SinEnc}(n, d)$ for block index $n$ is computed as follows:
\begin{equation}
    \text{SinEnc}(n, 2i) = \sin\left(\frac{n}{10000^{2i/d}}\right), \text{SinEnc}(n, 2i + 1) = \cos\left(\frac{n}{10000^{2i/d}}\right),
\end{equation}
where $i$ ranges from 0 to $\left\lfloor\frac{d-1}{2}\right\rfloor$. The sinusoidal encoding is then fed into embedding layers to obtain the block index embedding $emb\_n$.
% To effectively incorporate the block index into the weight generator as a new conditioning, we introduce an additional embedding mechanism for the block index $n$, which is similar to the embeddings of the timestep $t$. We first compute the sinusoidal block index encoding, which is represented as $[\sin, \cos]$ then feed the encoding to projection layers. 
Finally, the block index embedding $emb\_n$ is combined with the time step embedding $emd\_t$,  represented as $emb = emb\_n+emb\_t$, to be leveraged in each residual block in the generator. Thus, the weight generator has access to the block index throughout the denoising process. 
% The weight generator is based on the diffusion model architecture, which is composed of a text encoder, a UNet weight information creator, and an autoencoder decoder. To incorporate the block index into the weight generator, 
From the results, we observe that the block index $n$ can model the weights from different blocks effectively without the necessity to condition on prior predicted weights, while greatly reducing the computations.    

\subsection{Fast Fine-Tuning with Generated Weight Initializations} \label{sec:finetune}
When a new concept/style $T$ arises, the weight initializations can be obtained by conducting inference for the trained weight generator $\hat{\epsilon}_\theta$ for each weight block $n$. To achieve fast acquisition of weight initializations, we employ a direct reconstruction method to avoid the iterative denoising process. More specifically, at the selected time step t that leans to the noise side, we forward the denoising diffusion model to predict the
noise $\hat{\epsilon}_\theta(\mathbf{w}_n^t,t,n,\tau(T))$, and we conduct a direct recovery to obtain the real weight $\mathbf{w}_{n}=\mathbf{w}_{n}^0$:
\begin{equation}
 \mathbf{w}_{n}^0 = \frac{1}{\alpha_t} \mathbf{w}_{n}^t - \sigma_t \hat{\epsilon}_\theta(\mathbf{w}_n^t,t,n,\tau(T)).
\end{equation}
% In our experiments, we use DDIM to sample with the following iterative denoising process from $t$ to a previous time step $t'$, which can be represented as 
% \begin{equation}
%  \mathbf{w}_{n}^{t'} = \alpha_{t'} \frac{\mathbf{w}_{n}^t - \sigma_t \hat{\epsilon}_\theta(\mathbf{w}_n^t,t,n,\tau(T))} {\alpha_t} + \sigma_{t'} \hat{\epsilon}_\theta(\mathbf{w}_n^t,t,n,\tau(T)),
% \end{equation}
% where $\mathbf{w}_{n}^{t'} $ will be fed into $\hat{\epsilon}_\theta(\cdot)$ again until $t'$ becomes 0, i.e., the denoising process finishes.
After conducting inference for all of the $N$ weight blocks, we can obtain the weight initialization $\{\mathbf{w}_{n} \}_{n=1}^N$ for the concept/style $T$. To capture the details of the new concept/style better, a further fine-tuning process for the GAN weights is leveraged with the conditional GAN loss as follows 
\begin{equation} \label{eq:gan_loss}
\centering
\begin{aligned}
   &\min_{\mathbf{w}_{lora}} \max_{\mathbf{w}_d} \lambda \underbrace{ \mathbb{E}_{\mathbf{x},\tilde{\mathbf{x}}^T,\mathbf{z}, T} \left[ \| \tilde{\mathbf{x}}^T - \mathcal{G}(\mathbf{x}, \mathbf{z}, T;\mathbf{w}_g,\mathbf{w}_{lora}) \|_1 \right]}_{\textrm{$\ell_1$ loss}} + \\ &\underbrace{\mathbb{E}_{\mathbf{x}, \tilde{\mathbf{x}}^T} 	\left[\log \mathcal{D} (\mathbf{x},\tilde{\mathbf{x}}^T; \mathbf{w}_d) \right]  + \mathbb{E}_{\mathbf{x},{\mathbf{z}}, T} 	\left[\log (1- \mathcal{D} (\mathbf{x}, \mathcal{G}(\mathbf{x},\mathbf{z}, {T}; \mathbf{w}_g); \mathbf{w}_d)) \right]}_{\textrm{conditional GAN loss}},
   \end{aligned} 
\end{equation}
where $\tilde{\mathbf{x}}^T$ denotes images generated by the diffusion model conditioned on the concept $T$ of the target style, $\mathcal{G}$ is the generator with original weights $\mathbf{w}_g$ and the LoRA weights $\mathbf{w}_{lora}$, $\mathcal{D}$ denotes the discriminator function parameterized by $\mathbf{w}_d$, respectively, $\mathbf{z}$ is a random noise introduced to increase the stochasticity of output, and $\lambda$ can be used to adjust the relative importance between two loss terms. During the fine-tuning process, the generator only optimizes the LoRA weights $\mathbf{w}_{lora}$ which are initialized with the predictions $\{\mathbf{w}_{n} \}_{n=1}^N$.
By initializing the GAN weights from predictions, we are able to use 
% \textcolor{red}{consider that we claim efficient training, should we use a large learning rate? that sounds like training from scratch. small LR + fewer iterations?} 
much fewer training epochs to reach the same or even better FID performance. Besides fine-tuning after prediction, we also consider incorporating the GAN training loss in Eq. (\ref{eq:gan_loss}) to the weight prediction loss in Eq. (\ref{eq:diff}). However, through experiments, we find out that combining these two loss terms is not able to provide better performance, but leads to more computation costs for training the weight generator.  
\section{Experiments}

In this section, we provide the detailed experimental settings, results of our proposed method compared to baseline methods, and the ablation studies. More details as well as some ablation studies can be found in the Appendix. 
\subsection{Experiment Settings}
\begin{figure}[t]
     \centering
     \includegraphics[width=0.8\linewidth]{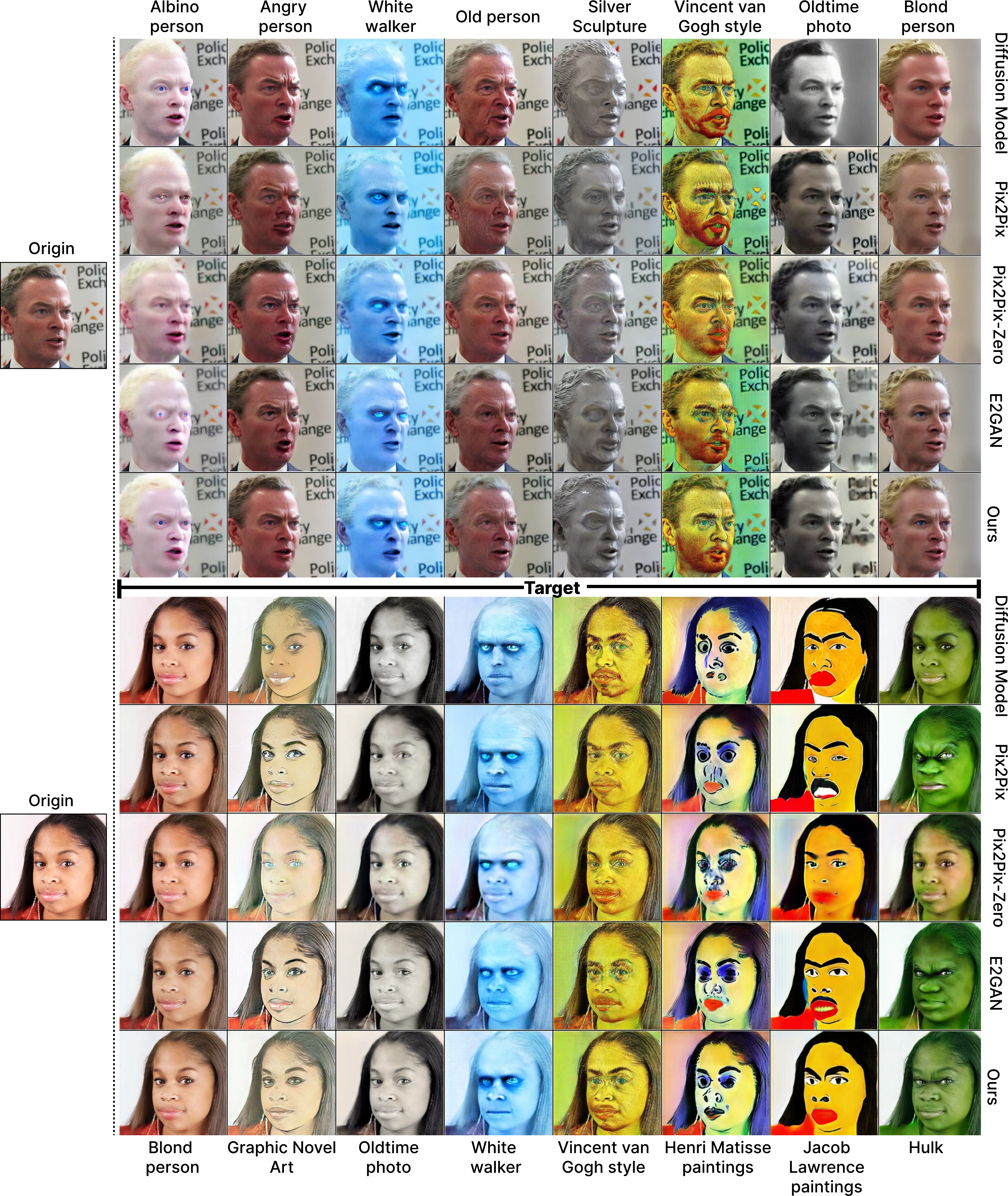}  
     \caption{{\textbf{Qualitative comparisons} across different concept domains. The \emph{leftmost} column shows two original images and the remaining columns present the corresponding synthesized images in the target concept domain, where target prompts are shown at the top/bottom row. We provide images generated by various models.} }
    \label{fig:main_qualitive}  
    \vspace{-4mm}
\end{figure}
\subsubsection{Baselines.} We compare our method with image-to-image translation methods like pix2pix~\cite{isola2017image} (image generator with $9$ ResNet blocks), pix2pix-zero-distilled that distills Co-Mod-GAN~\cite{zhao2021large} from pix2pix-zero~\cite{parmar2023zero}, and efficient GAN training methods E$^2$GAN \cite{gong20242}.

\subsubsection{Prompt Dataset Preparation.}
We first use ChatGPT-3.5 \cite{brown2020language} to collect prompts for the three categories: 1) art, 2) characteristic, and 3) facial modification concepts as discussed in Sec. \ref{sec:data_collection}. After filtering out repeated and unmeaning prompts, we get 226 art concepts, 441 character concepts, and 26 facial modification concepts. We reserve 20 art concepts, 20 character concepts, and 5 facial modification concepts for test use, never used during the weight generator training. By combining the concepts across different categories that are not reserved as test concepts and filtering, the prompt dataset is enriched with another 84477 concepts. We further augment the obtained concepts with Vicuna \cite{chiang2023vicuna} for concepts with the same meaning but different expressions and filter meaningless ones, which leads to an additional 4126 augmented art concepts, 8070 augmented characteristic concepts, and 245 augmented facial modification concepts.  

\subsubsection{Paired Image Preparation.} After the prompt dataset is collected, we generate images for GAN training for each concept. We verify our method on $1,000$ images from the FFHQ dataset~\cite{karras2019style} with image resolution as $256\times256$. The images in the target domain are generated with several different text-to-image diffusion models, including Stable Diffusion~\cite{rombach2022high}, Instruct-Pix2Pix~\cite{brooks2022instructpix2pix},  Null-Text Inversion~\cite{mokady2022null}, ControlNet~\cite{zhang2023adding}, and InstructDiffusion~\cite{geng2023instructdiffusion}. The generated images with the best perceptual quality among diffusion models are selected to form the real images into paired datasets. To perform training and evaluation of GAN models, we divide the image pairs from each target concept into training/validation/test subsets with the ratio as $80\%$/$10\%$/$10\%$. 

\subsubsection{Ground-truth GAN Weights Preparation.} With the paired images, we collect the ground-truth GAN weights to train the weight generator. We apply LoRA to each layer of the generator, leading to 0.2256M weights for each concept to learn. We follow the standard approach that alternatively updates the generator and discriminator~\cite {goodfellow2020generative}.
The training is conducted from an initial learning rate of 2e-4 with mini-batch SGD using Adam solver~\cite{kingma2014adam}.
% The training is conducted with 200 epochs.
The total training epochs are set to $100$. The obtained weights are grouped following Sec. \ref{sec:data_format} and divided with a block size of 256, resulting in a total of 854 weight blocks. When dividing the weights into equal-sized blocks, zeros are padded when necessary. 

\subsubsection{Training Settings.} The weight generator is trained with AdamW optimizer \cite{kingma2014adam}. The initial learning rate is set as 1e-5, the weight decay is set as 0.01, and the training batch size is set as 512. The weight generator training is conducted with 4 or 8 nodes, each with 8 NVIDIA A100 GPUs with 40GB or 80GB memory. The block size for weight division is set as 256. For the following fine-tuning process, to show a fair comparison with the efficient GAN training approach E$^2$GAN, we adopt the same cluster size as 400 for each concept. The initial learning rate is set as 0.0015 and the fine-tuning epochs are set as 20.

\subsubsection{Evaluation Metric.} We compare the performance of our efficient weight generalization by comparing images generated by models obtained via our approach and baseline methods. The evaluation is achieved by calculating Clean FID proposed by~\cite{parmar2022aliased} on the test sets of the paired images. 

\subsection{Experimental Results}\label{sec:overall_performance}
% \begin{table}[htb]
% \centering \small
% \caption{The FID of GAN-Adapter compared with traditional GAN training on various style.}
% \begin{tabular}{l|c|ccc}
% \toprule
% \textbf{Method} & \multicolumn{1}{l|}{\textbf{Traditional train}} & \multicolumn{3}{c}{\textbf{GAN-Adapter}} \\  \midrule
% Train param num & 7.841M & \multicolumn{3}{c}{0.777M} \\  \midrule
% \diagbox{ Target style $s_t$}{Original style $s_o$}& - & Old person & Young person & Blond \\  \midrule
% Old person & 73.52 & - & 69.80 & \textbf{69.59} \\
% Young person & 56.22 & 52.64 & - & \textbf{52.16} \\
% Blond & 70.41 & \textbf{62.58} & 63.22 & - \\
% Grey hair & \textbf{67.61} & 71.60 & 74.55 & 73.99 \\
% Angry & 57.75 & \textbf{52.11} & 53.56 & 54.48 \\
% Pale & 50.96 & 45.65 & 48.14 & \textbf{45.44} \\
% Tan & 52.58 & 47.23 & 47.89 & \textbf{46.95} \\
% Zombie & \textbf{70.01} & 71.70 & 76.67 & 75.50 \\
% Curly &{67.47} & \textbf{64.79} & 65.55 & 64.85 \\ \bottomrule
% \end{tabular}
% \end{table}

% To demonstrate the effectiveness of~E$^2$GAN, we show both qualitative and quantitative performance. 

\subsubsection{Qualitative Results.}
The synthesized images in the target domain obtained by our method and other methods are shown in Fig.~\ref{fig:main_qualitive}. The original images are listed in the leftmost column, and the synthesized images for the target concept obtained by diffusion models, pix2pix, pix2pix-zero-distilled, E$^2$GAN, and ours are shown from top to bottom.
The tasks span a wide range, such as changing the age, artistic styles, and characteristic styles.
According to the results, the models obtained by ours can modify the original images to the target concept domain by fast fine-tuning with the weight initializations from the weight generator. For instance, for the \texttt{Jacob Lawrence paintings} prompt on the FFHQ dataset, our model generates more meaningful images compared to all baseline methods. As for the \texttt{albino person} prompt, our method edits the image as desired while having fewer artifacts. We provide more qualitative results in the Appendix.

\subsubsection{Quantitative Results.}
% \begin{table}[]
\begin{wraptable}{r}{0.62 \textwidth}
\vspace{-10mm}
\centering
\caption{ \textbf{FID and time consumption comparison.} FID is calculated between the images generated by GAN-based approaches and diffusion models.
Reported FID is averaged across different concepts in the test prompt dataset.} \label{tab:comparison}
\scalebox{0.9}{
\begin{tabular}{l|ccc}
% \toprule
\Xhline{0.2ex}

% Train param.  & 7.841M & \multicolumn{3}{c}{0.777M} \\  \hline & AFHQ  & 78.91 & 80.92 & 79.01 
% \diagbox{\textbf{Method}}{\textbf{Dataset}} &   FFHQ   & Landscape \\ \hline 
\textbf{Method} &   \textbf{FID}   & \textbf{Time Consumption} \\ \hline 
Pix2pix-zero-distilled & 144.81  & 112 min    \\                    
Pix2pix                & 99.20  &   659.8 secs  \\
E$^{2}$GAN & 93.86  & 198.5 secs      \\
\rowcolor[gray]{.9} \textbf{Ours}             & \textbf{89.93} & \textbf{43.3 secs} \\
\Xhline{0.2ex}
% \bottomrule
\end{tabular}
}
\vspace{-4mm}
% \end{table}
\end{wraptable}
We compare the quantitive results and training time consumption between our method and other baseline methods, and the results are provided in Tab. \ref{tab:comparison}. Note that for each concept, pix2pix-zero-distilled and pix2pix are trained on the whole training dataset of $800$ samples with 200 epochs. E$^2$GAN begins with a base model and is fine-tuned with $400$ samples for 100 epochs. Our method initializes the GAN weights with the prediction from the weight generator and is fine-tuned with $400$ samples for 20 epochs.  The reported FID values are computed with an average for all the concepts in the test prompt dataset. The time consumption is measured on one NVIDIA H100 GPU. 
The results demonstrate that our method can reach an even better FID performance than the conventional GAN training techniques like pix2pix and pix2pix-zero-distilled, and efficient GAN training methods such as E$^2$GAN, indicating the high-fidelity of generated images. Furthermore, our method greatly reduces the time consumption for obtaining the model for a new concept/style. The time consumption of our method is mainly composed of two parts: 1) the inference of the weight generator to get the predicted weight initialization values for each block; and 2) the fine-tuning process with the initialized weight values. By using a batch size of 64, \textbf{the prediction process of the weight generator only takes 1.19 secs}. Taking advantage of the effective weight initializations, the fine-tuning process only requires 20 epochs to reach good performance, which takes 42.1 secs and results in 43.3 secs of training time in total. Specifically, our method reduces the training time by 152$\times$, 15$\times$, and 4.6$\times$ compared to Pix2pix-zero-distilled, Pix2pix, and E$^2$GAN, respectively. Furthermore, the results also indicate the effectiveness of only updating LoRA weights to transfer the generative domain of the GAN model as both our method and E$^2$GAN only update LoRA weights, but achieve good FID performance. 

\begin{figure}[t]
     \centering
     \includegraphics[width=0.9\columnwidth]{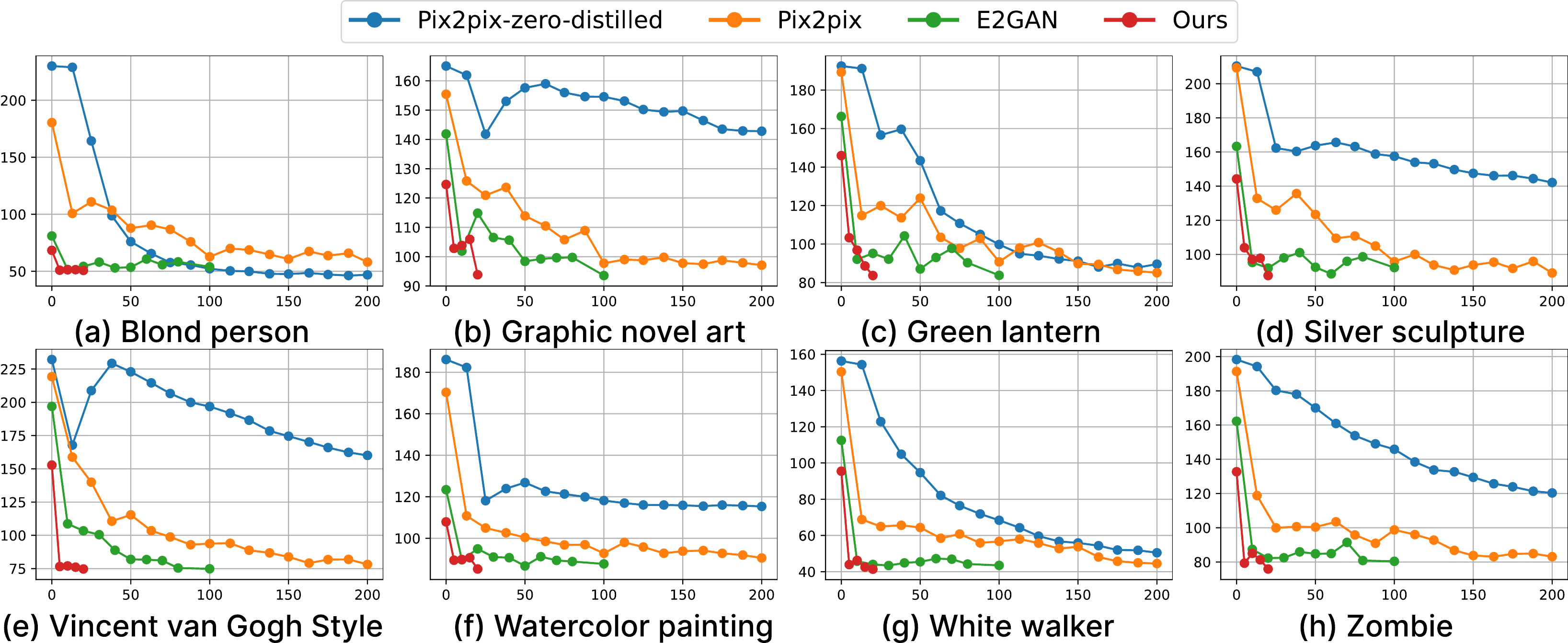}  
     \caption{The FID performance comparison between our method and baseline methods along with the training process on the test dataset for different concepts/styles. 
     % Each sub-figure represents the training curve for one concept, indicated by the title at the bottom. Our method is able to provide a better initialization with better FID before training, and a faster convergence.  
     }
    \label{fig:fid_curve}  
\end{figure}

\subsubsection{FID Curves.} We show the FID curves along with the training process by different methods in Fig. \ref{fig:fid_curve}. More FID curve results can be found in the Appendix. From the results, we can observe that our method achieves a faster convergence across different concepts/styles. The effectiveness of our weight generator for generating weight initializations is revealed in two aspects. First, our method provides a better starting point for the training, indicated by the better FID score before training (epoch 0). Second, training from the initialization can reach a better FID score with much fewer training epochs. For instance, for the concept \texttt{white walker}, our weight generator provides a weight initialization with an FID of 95.46, while the baseline methods have an initial FID of 154.29, 150.32, and 112.50, respectively. After fine-tuning with 20 epochs, our method reaches an FID of 41.38, which improves the baseline methods by at least 2.1.
% The efficient GAN training approach E$^2$GAN also has a better starting FID score than conventional GAN training approaches. The rationale behind this is that E$^2$GAN starts from a pre-trained base model on multiple concepts. 

\subsubsection{Ablation Studies.} We conduct ablation studies on the \textbf{block size} and \textbf{weight grouping} rules. Due to the huge training cost of the weight generator on the entire training dataset, we conduct small-scale experiments for the ablation study. We overfit the weight generator solely on the weights corresponding to a particular concept of interest. The approach provides a precise assessment of how well a particular configuration captures a specific concept. Any discrepancies between the overfitted results and the ground-truth values can be attributed directly to the efficacy of the chosen configuration. 

% \begin{table}[]
\begin{wraptable}{r}{0.63 \textwidth}
\vspace{-12mm}
\centering
\caption{ \textbf{Ablation study on block size} for weight division in data preparation. The first row represents the performance of the ground-truth trained model while the remaining rows correspond to the performance of generated weights from the weight generator with different weight block sizes.} \label{tab:ablation_blk_size}
\scalebox{0.9}{
\begin{tabular}{c|c|cc}
\Xhline{0.2ex}
\multirow{2}{*}{\textbf{Block Size}} & \multirow{2}{*}{\textbf{Weight Gene. Time}} & \multicolumn{2}{c}{\textbf{FID}}                 \\ \cline{3-4} 
                            &                                         & \multicolumn{1}{c|}{\texttt{Grey hair}} & \texttt{Batik}  \\ \hline
-                           & -                                       & \multicolumn{1}{c|}{89.04}     & 90.43  \\
128                         & 1.71 secs                               & \multicolumn{1}{c|}{107.67}    & 128.44 \\
\rowcolor[gray]{.9}   \textbf{256}                         & 1.19 secs                               & \multicolumn{1}{c|}{\textbf{98.35}}     & \textbf{94.36}  \\
512                         & 1.04 secs                               & \multicolumn{1}{c|}{130.85}    & 118.31 \\ \Xhline{0.2ex}
\end{tabular}
}
\vspace{-4mm}
% \end{table}
\end{wraptable}
For the block size study, we investigate different block size settings including 128, 256, and 512, on two randomly selected concepts \texttt{grey hair} and \texttt{Batik}. The block size selection is based on the size of all layers in the GAN model. The results are shown in Tab. \ref{tab:ablation_blk_size}. The results show that the block size selection has an impact on the weight generation performance. Setting a larger block size leads to a faster weight generation process. However, the FID performance is the best when the block size is set as 256, while the generation time is slightly slower than a block size of 512. The results indicate that the appropriate selection of the block size to divide grouped weights is important for achieving good performance of the weight generator. 

% \begin{table}[]
\begin{wraptable}{r}{0.42 \textwidth}
\vspace{-10mm}
\centering
\caption{ \textbf{Ablation study on weight grouping.} } \label{tab:ablation_group}
\scalebox{1}{
\begin{tabular}{c|cc}
 \Xhline{0.2ex}
\textbf{Group Rule} & \texttt{Grey hair} & \texttt{Batik}  \\ \hline
\rowcolor[gray]{.9}  \textbf{Rule 1)}    & \textbf{98.35}     & \textbf{94.36}  \\
Rule 2)    & 98.40     & 95.17  \\
Rule 3)    & 122.71    & 126.50 \\  \Xhline{0.2ex}
\end{tabular}
}
\vspace{-4mm}
% \end{table}
\end{wraptable}
For the weight grouping before weight division, we study 3 different rules including 1) group the LoRA down layer, LoRA up layer, and the following BN layer if applicable for each layer $i$ to one group, and append the reshaped 1-dimensional weight vectors one by one; 2) group the LoRA down layer, LoRA up layer, and the following BN layer if applicable for each layer $i$ to one group, concatenate the weights through the channel dimension and reshape it to 1-dimensional vector; 3) view each layer as a single group and reshape to 1-dimensional vector. Besides the weight grouping rules, all the other settings are the same as the default setting for the main results. We show the comparisons of the 3 rules in Tab. \ref{tab:ablation_group}. From the results, we can see that Rule 1) and Rule 2) both perform much better than Rule 3), which indicates the importance of grouping the layers belonging to the same layer $i$ into one group. The rationale behind this might be related to the different statistics among different layers. Furthermore, Rule 1) and Rule 2) do not have obvious performance differences, which means after grouping weights for the same layer $i$ together, it is not necessary to take different channels separately. Combining the ablation studies, we get the default setting used in the main results, which correspond to use rule 1) to group weights and set the block size as 256.

\section{Conclusion}
This paper studies to generate good weight initializations with a weight generator to reduce the training cost of a DNN. Leveraging the image-to-image translation task with GANs as a case study, we demonstrate the feasibility and effectiveness of our approach.  Through the division of weights into equal-sized blocks and the incorporation of block indexes, we mitigate the complexity of varied layer characteristics and a large number of weights. By training a diffusion process with both textual concept conditions and block indexes, the weight generator produces weight initializations for new concepts/styles efficiently with a one-step direct recovery. We conduct extensive experiments on different concepts to demonstrate the effectiveness of our proposed framework. By leveraging the synthesized weight initializations, we can achieve better FID performance with much fewer training costs across various concepts/styles than baseline methods including conventional GAN training and efficient GAN training approaches.
We reduce the time consumption for obtaining the model for a new concept by 4.6$\times$  while improving the FID performance by 3.93 than efficient GAN training baseline and reduce the total training time by $15\times$ than training from scratch  (\ie, Pix2pix~\cite{isola2017image}) with better FID performance.

\section{Discussion of limitations}
 The development of a weight generator to synthesize improved weight initializations can increase the efficiency and efficacy of model training. 
  To prepare the training data for the weight generator, we leverage diffusion models to edit real images, thereby obtaining edited images that encompass a wide range of concepts. This approach allows us to create paired data spanning various concepts/styles, which provide the foundation for training diverse GAN weights for different generation domains. However, the quality of the generated images plays a pivotal role in influencing the performance of the trained GAN model, consequently impacting the performance of the weight generator.
  % To prepare the training data for the weight generator, we use diffusion models to edit real images to obtain the edited images, forming pairs of data across different concepts/styles to train different GAN weights. The generated image quality influences the trained GAN model in achieving the target concept/styles and further influences the performance of the weight generator. 
  While diffusion models offer a powerful tool for image editing, the quality and fidelity of the generated images may not always meet the desired standards. Furthermore, utilizing diffusion models for data collection remains expensive. Developing efficient techniques
to rapidly construct well-paired and high-quality images
from diffusion models would greatly enhance the training of the weight generator.

\clearpage
\setcounter{page}{1}
\appendix

\begin{nolinenumbers}
\maketitlesupplementary
\end{nolinenumbers}
% \clearpage
% \newpage
\setcounter{section}{0}
\setcounter{figure}{0}
\setcounter{equation}{0}
\makeatletter 
\renewcommand{\thefigure}{A\@arabic\c@figure}
\makeatother
\setcounter{table}{0}
\renewcommand{\thetable}{A\arabic{table}}
\renewcommand{\theequation}{S\arabic{equation}}

\appendix
% \section*{Appendix}

\section{More Implementation Details}
% \textbf{Details for Diffusion Model.}
\subsection{Details for Diffusion Model}
We apply the most recent diffusion-based image editing models to create paired datasets, which include Stable Diffusion (SD)~\cite{rombach2022high}, Instruct-Pix2Pix (IP2P)~\cite{brooks2022instructpix2pix},  Null-text inversion (NI)~\cite{mokady2022null}, ControlNet~\cite{zhang2023adding}, and Instruct Diffusion \cite{geng2023instructdiffusion}. 
For all these models, we use the checkpoints or pre-trained weights reported from their official websites\footnote{SD v1.5:~\url{https://huggingface.co/runwayml/stable-diffusion-v1-5}, IP2P:~\url{http://instruct-pix2pix.eecs.berkeley.edu/instruct-pix2pix-00-22000.ckp}, NI:~\url{https://huggingface.co/CompVis/stable-diffusion-v1-4}, ControlNet:~\url{https://huggingface.co/lllyasviel/ControlNet/blob/main/models/control_sd15_normal.pth}, InstructDiffusion:~\url{https://github.com/cientgu/InstructDiffusion}.}.

More specifically, for SD, the strength, guidance scale, and denoising steps are set to $0.68$, $7.5$, and $50$, respectively. 
For IP2P, 
% we use the pretrained checkpoint from \href{http://instruct-pix2pix.eecs.berkeley.edu/instruct-pix2pix-00-22000.ckpt}{[link]}. The 
images are generated with $100$ denoising steps using a text guidance of $7.5$ and an image guidance of $1.5$.
For NI, each image is generated with $50$ denoising steps and the guidance scale is $7.5$. The fraction of steps to replace the self-attention maps is set in the range from $0.5$ to $0.8$ while the fraction to replace the cross-attention maps is $0.8$. The amplification value for words is $2$ or $5$, depending on the quality of the generation. For ControlNet, 
the control strength, normal background threshold, denoising steps, and guidance scale are $1$, $0.4$, $20$, and $9$, respectively. For Instruct Diffusion, the denoising steps, text guidance, and image guidance are set as $100$, $5.0$, and $1.25$, respectively.  

% \textbf{Hyperparameters in LoRA Rank Search.}
% \subsection{Data Challenge}

\subsection{LoRA Rank}
We conduct a similar LoRA search process as in \cite{gong20242}. During the process of searching LoRA rank, the rank $r_i$ for each layer $i$ is upscaled once for every $e$ epoch until $r_i$ reaches the upper threshold $\tau_i$ for the layer $i$. In the experiments, $e$ is set as $10$. The rank threshold $\tau_i$ is determined by the size of the layer. More specifically, the layers include (1) four CONV-based upsampling layers with the shape as ${[3,64,7,7],[64,128,3,3],[128,256,3,3]}$, and ${[256,256,3,3]}$; (2) four corresponding downsampling layers by transpose CONV with the same set of weight shape as upsampling; (3) transformer blocks with projection matrices $q$,$k$,$v$ with shape as ${[256,256]}$, and multi-layer perceptron (MLP) module with shape as ${[2048,256]}$ and ${[256,1024]}$; and (4) ResNet blocks with CONV layers with the shape as $[256, 256, 3, 3]$. Based on the weight size, the rank threshold $\tau$ is set as $1$, $4$, $16$, and $32$ for the four upsampling/downsampling layers, respectively, $1$ for the layers in the transformer block, and $32$ for the CONV layers in the ResNet block. After the search process, the suitable rank is determined as $1$, $4$, $8$, and $8$ for the four upsampling/downsampling layers, and $8$ for the CONV layers in the ResNet blocks.

\section{Examples of Collected Prompts}
\begin{figure}[htbp]
     \centering
     \includegraphics[width=0.8\columnwidth]{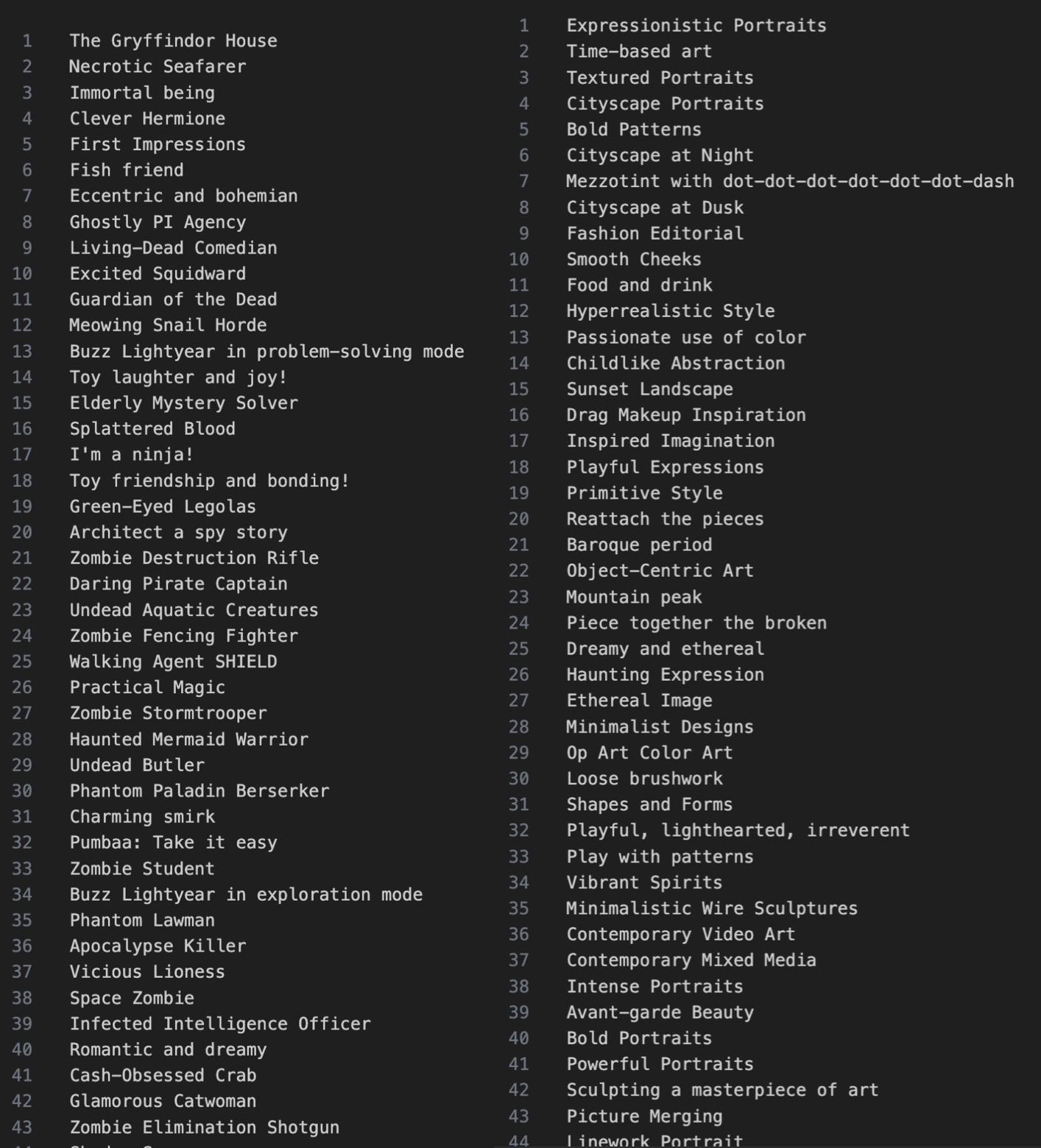}  
     \caption{Examples of collected text prompts of concepts/styles. 
     % Each sub-figure represents the training curve for one concept, indicated by the title at the bottom. Our method is able to provide a better initialization with better FID before training, and a faster convergence.  
     }
    \label{fig:prompt_example}  
\end{figure}

We show the examples of the collected prompts of the concepts/styles in Fig. \ref{fig:prompt_example}. The prompts are generated by querying a large language model (Vicuna \cite{chiang2023vicuna}) to augment the prompt with the same meanings but different expressions.

% \section{More Explanations for the Block Size}
% 
\begin{figure}[t]
     \centering
     \includegraphics[width=0.7\columnwidth]{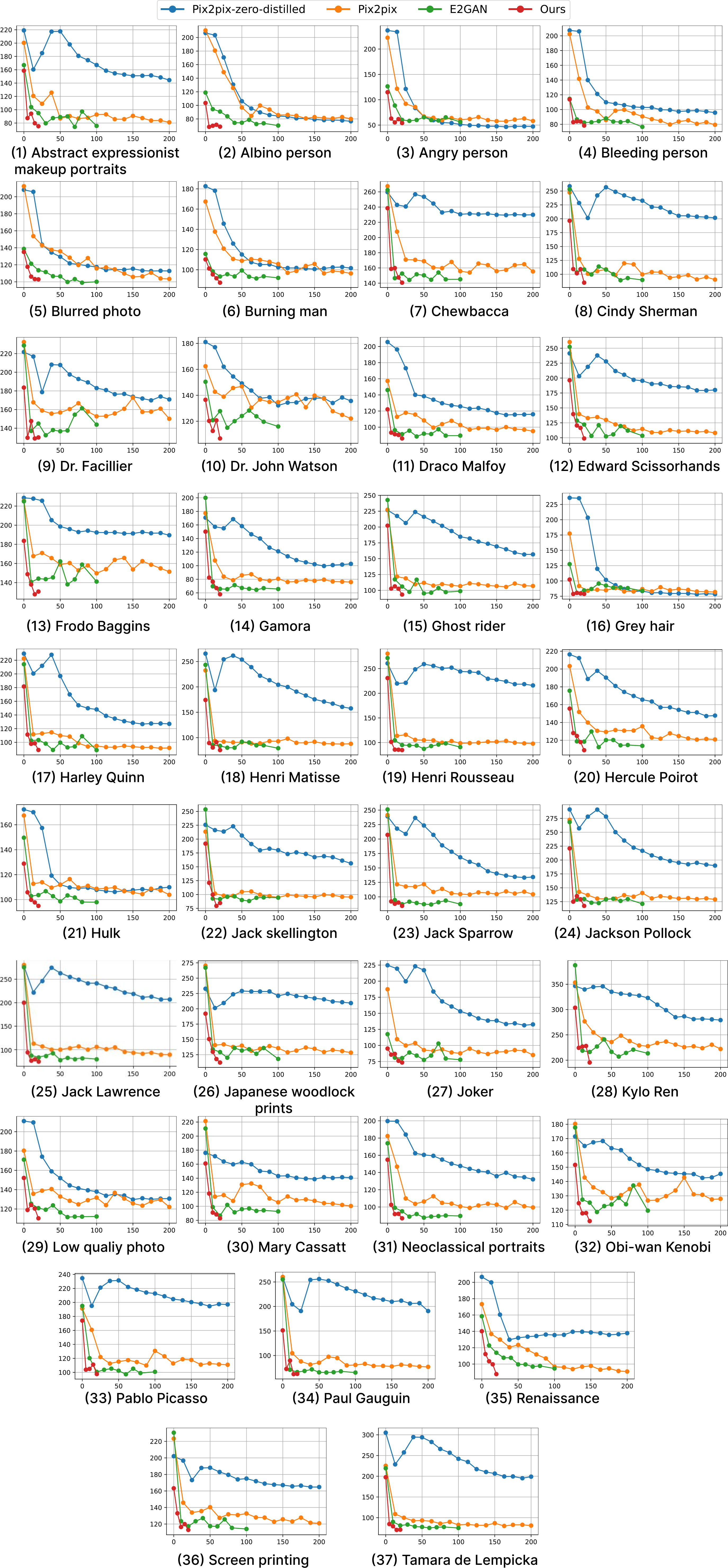}  
     \caption{Additional FID performance comparison between our method and baseline methods along with the training process on the test dataset for different concepts/styles. 
     % Each sub-figure represents the training curve for one concept, indicated by the title at the bottom. Our method is able to provide a better initialization with better FID before training, and a faster convergence.  
     }
    \label{fig:appendix_fid_curve}  
\end{figure}

\section{Additional FID Curves}
We provide the additional FID curves along with the training process obtained by different methods in Fig. \ref{fig:appendix_fid_curve}. According to the results, we can observe that our method achieves a faster convergence across different concepts/styles, which demonstrates the effectiveness of our weight generator for generating weight initializations. Meanwhile, the weight initializations generated by the weight generator reach a better FID performance before the training starts. 

\begin{figure}[t]
     \centering
     \includegraphics[width=1\linewidth]{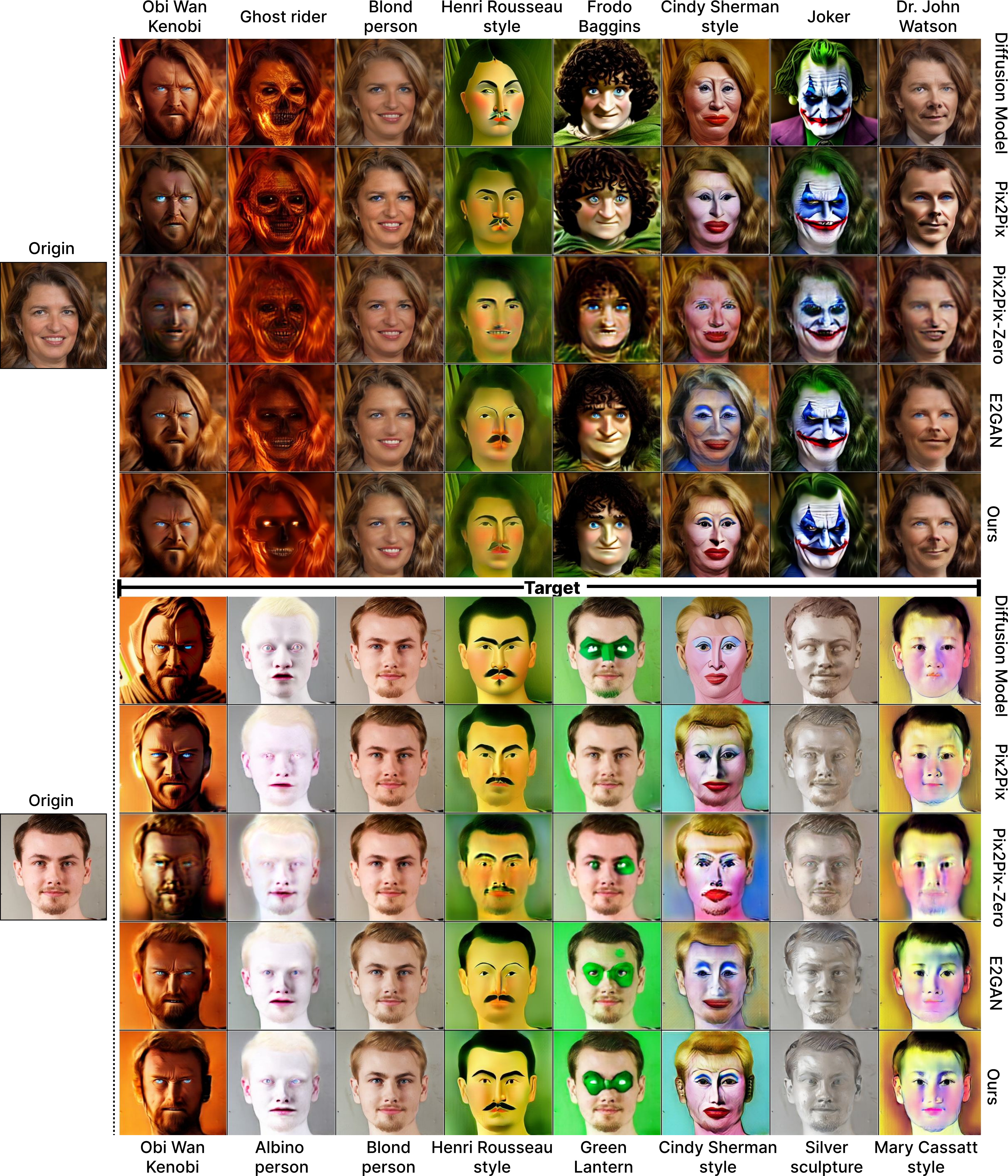}  
     \caption{{\textbf{Qualitative comparisons} across different concept domains. The \emph{leftmost} column shows two original images and the remaining columns present the corresponding synthesized images in the target concept domain, where target prompts are shown at the top/bottom row. We provide images generated by various models.} }
    \label{fig:appendix_1}  

\end{figure}

\begin{figure}[t]
     \centering
     \includegraphics[width=1\linewidth]{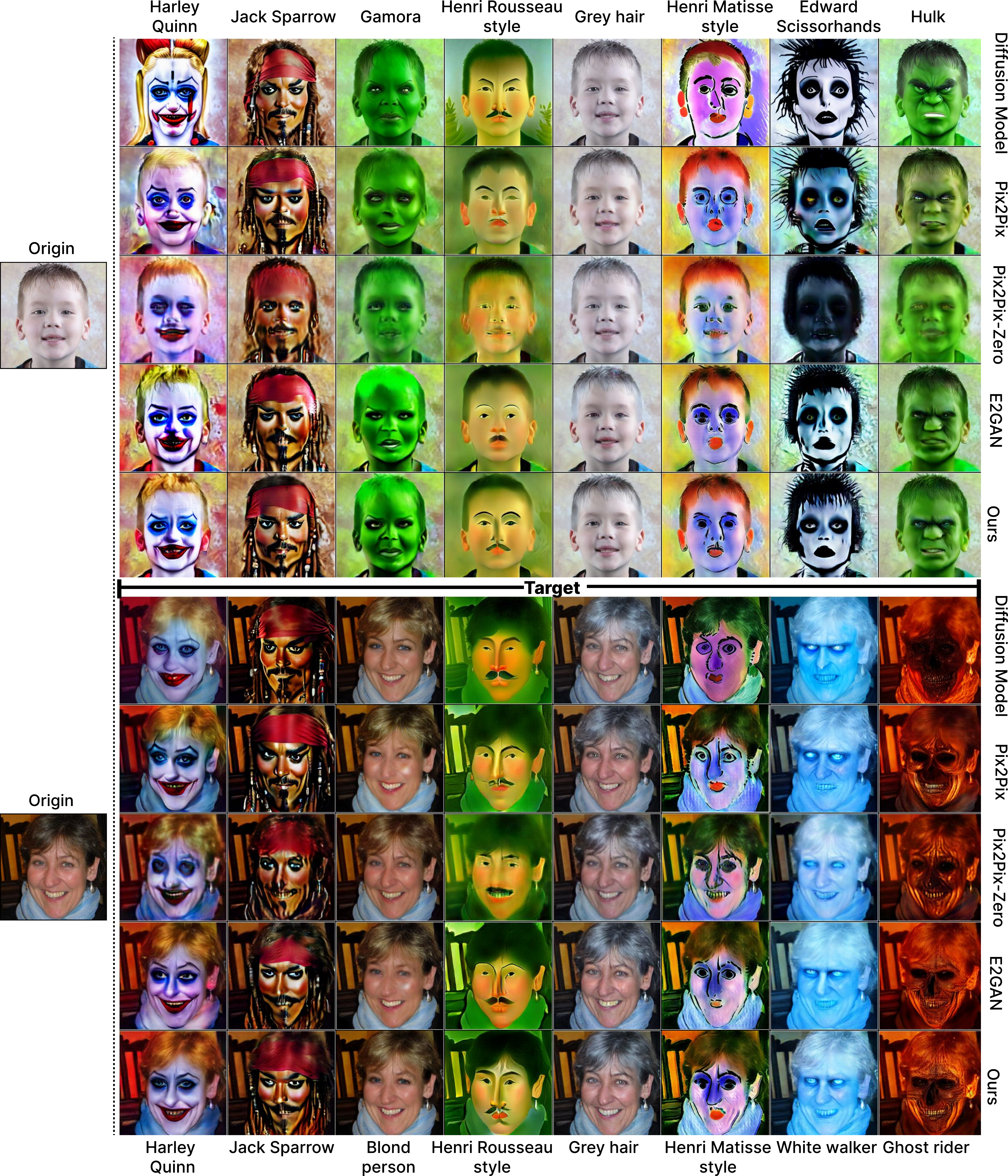}  
     \caption{{\textbf{Qualitative comparisons} across different concept domains. The \emph{leftmost} column shows two original images and the remaining columns present the corresponding synthesized images in the target concept domain, where target prompts are shown at the top/bottom row. We provide images generated by various models.} }
    \label{fig:appendix_2}  

\end{figure}

\begin{figure}[t]
     \centering
     \includegraphics[width=1\linewidth]{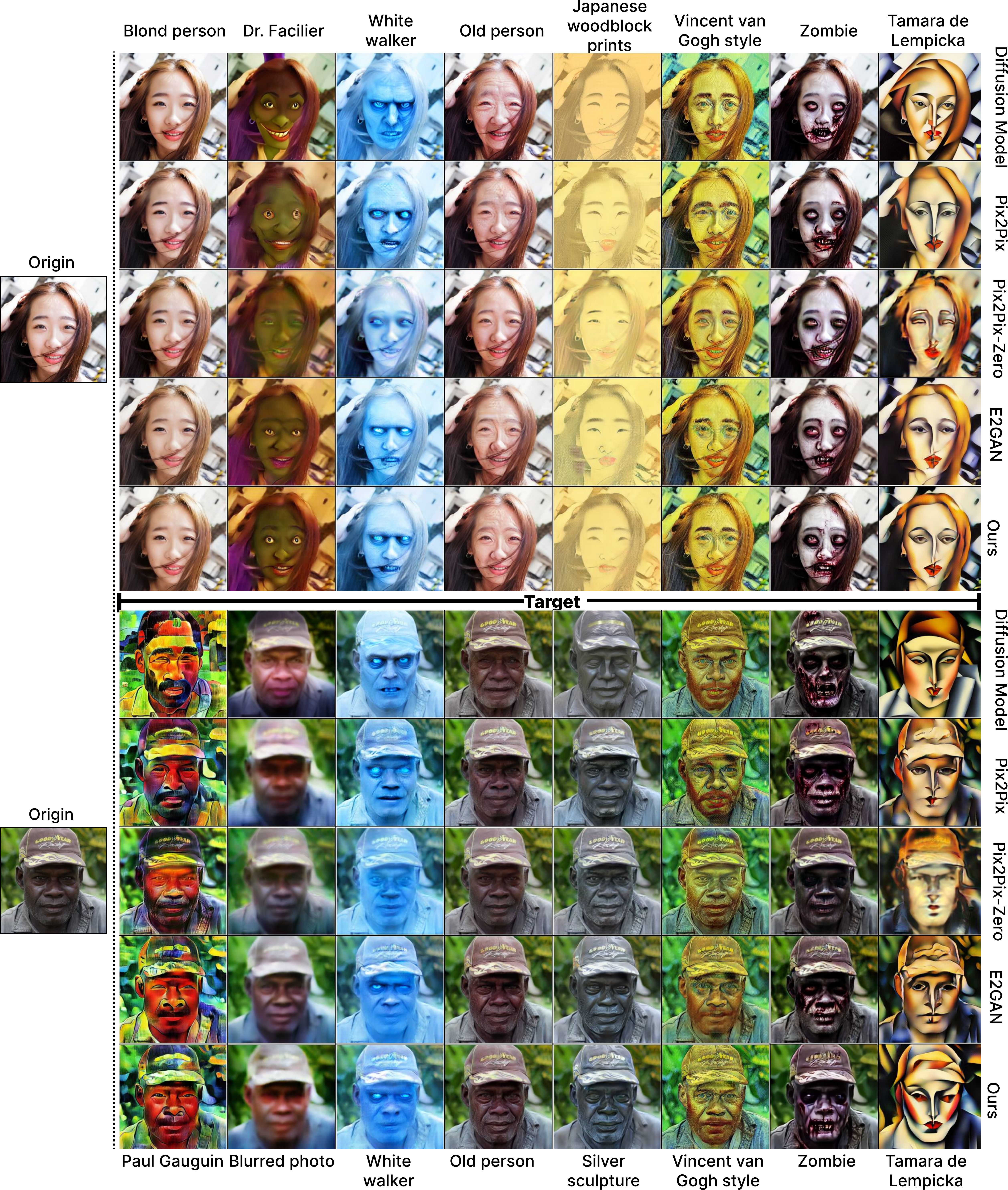}  
     \caption{{\textbf{Qualitative comparisons} across different concept domains. The \emph{leftmost} column shows two original images and the remaining columns present the corresponding synthesized images in the target concept domain, where target prompts are shown at the top/bottom row. We provide images generated by various models.} }
    \label{fig:appendix_3}  
\end{figure}

\begin{figure}[t]
     \centering
     \includegraphics[width=1\linewidth]{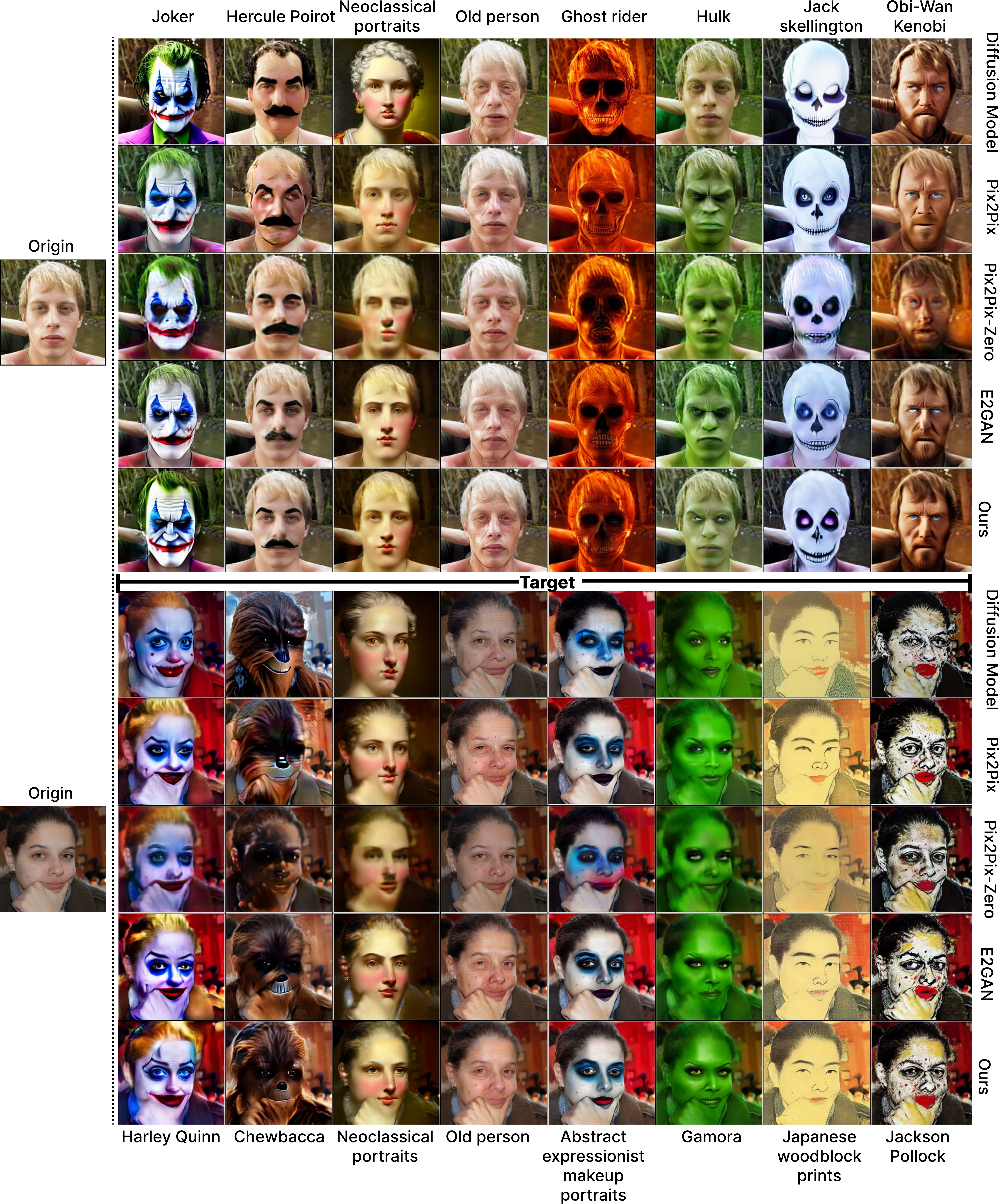}  
     \caption{{\textbf{Qualitative comparisons} across different concept domains. The \emph{leftmost} column shows two original images and the remaining columns present the corresponding synthesized images in the target concept domain, where target prompts are shown at the top/bottom row. We provide images generated by various models.} }
    \label{fig:appendix_4}  
\end{figure}

\begin{figure}[t]
     \centering
     \includegraphics[width=1\linewidth]{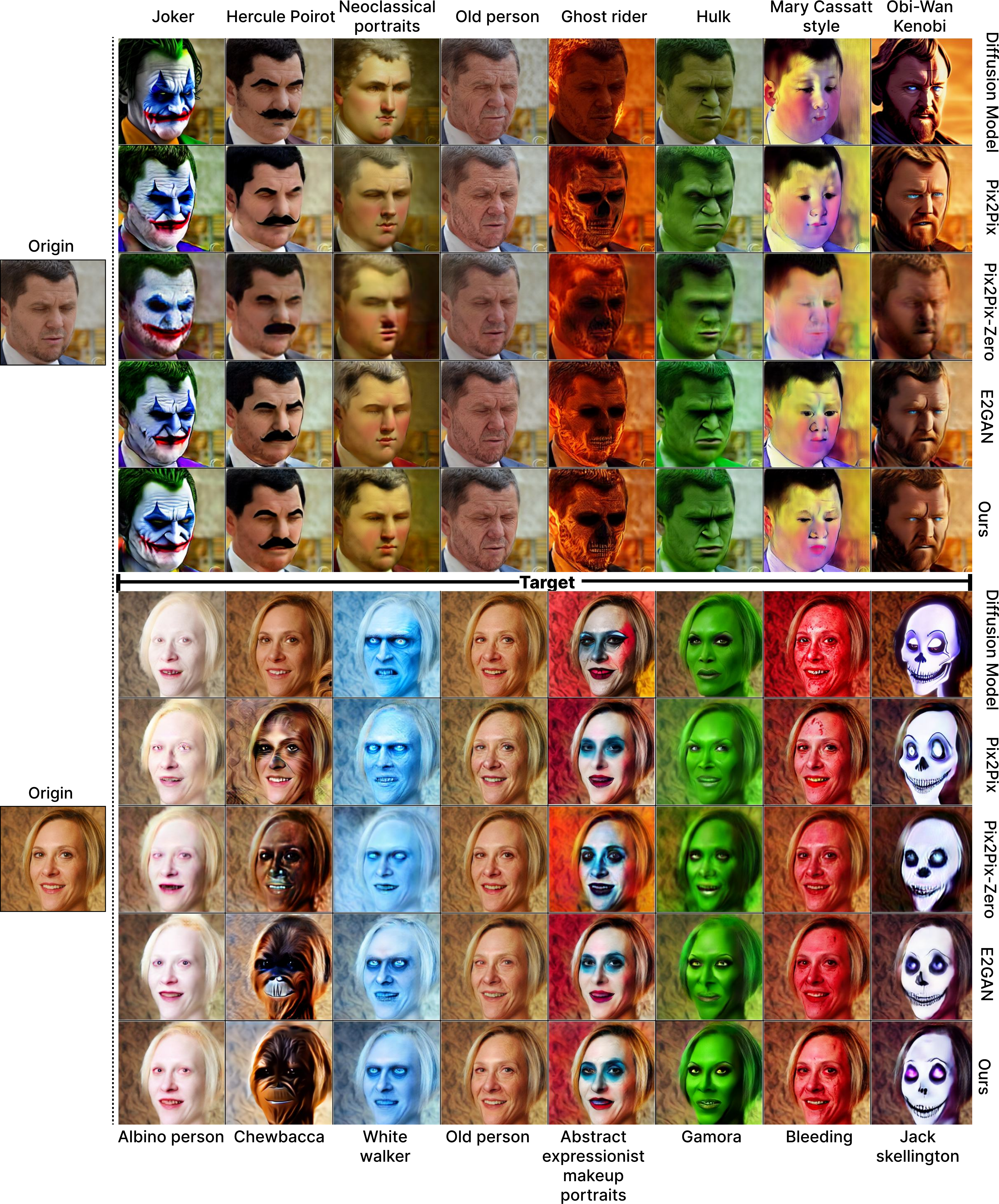}  
     \caption{{\textbf{Qualitative comparisons} across different concept domains. The \emph{leftmost} column shows two original images and the remaining columns present the corresponding synthesized images in the target concept domain, where target prompts are shown at the top/bottom row. We provide images generated by various models.} }
    \label{fig:appendix_5}  
\end{figure}
\section{Additional Qualitative Results}
We provide more example images generated by the models trained from the initializations generated by our weight generator and other baseline methods in 
Fig. \ref{fig:appendix_1}, \ref{fig:appendix_2}, \ref{fig:appendix_3}, \ref{fig:appendix_4}, \ref{fig:appendix_5}.

\section{End-to-End Training}
\begin{table}[htbp]
% \small
\centering
\caption{The FID comparison between noise prediction training and end-to-end training.} \label{tab:e2e}
\scalebox{0.95}{
\begin{tabular}{c|c|c}
\Xhline{0.2ex}
\textbf{Concept}              & \cellcolor[gray]{.9} \begin{tabular}[c]{@{}c@{}}     
      \textbf{\textbf{Noise prediction}} \\ 
      \textbf{\textbf{training}} 
      \end{tabular} & \begin{tabular}[c]{@{}c@{}}     
      \textbf{\textbf{End-to-end}} \\ 
      \textbf{\textbf{training}} 
      \end{tabular} \\ \hline
\texttt{Grey hair}   & \cellcolor[gray]{.9} \textbf{76.52}          & 77.62         \\
\texttt{Zombie}        & \cellcolor[gray]{.9} \textbf{79.33}          & 81.75          \\
\texttt{Amrita Sher Gil} & \cellcolor[gray]{.9} \textbf{95.14}          & 103.87          \\ \Xhline{0.2ex}
\end{tabular}}
\end{table}
With the noise prediction problem defined in Eq. \ref{eq:diff} and the fine-tuning objective for the GAN model with conditional GAN loss in Eq. \ref{eq:gan_loss}, one may wonder if we could directly train the weight generator with an end-to-end loss by combining these two loss terms. We conduct an ablation study to investigate the end-to-end training case. Due to the huge training cost of the weight generator on the entire training dataset, we conduct small-scale experiments for the ablation study. We overfit the weight generator solely on the last LoRA up layer in the generator with several concepts/styles. For end-to-end training, we assign a hyperparameter $\lambda$ as 0.01 with the conditional GAN loss to combine the two terms. When calculating the conditional GAN loss, we train the GAN model with 10 randomly sampled image pairs. We directly test the predicted overfitting results by plugging in the predicted weight values and testing the FID performance without further fine-tuning. We show the results in Tab. \ref{tab:e2e}. From the results, we can observe that end-to-end training does not improve the performance compared to solely using the noise prediction training loss but increases the training cost of the weight generator. Thus, we only adopt the noise prediction loss for the weight generator training.

% \clearpage  % TODO REVIEW/FINAL: This \clearpage needs to be removed from both review and camera-ready versions.

\clearpage

\bibliographystyle{splncs04}
\bibliography{main}
\end{document}